\documentclass[11pt,a4paper]{article}
\usepackage{emnlp-ijcnlp-2019}
\usepackage{times}
\usepackage{latexsym}
\usepackage{CJKutf8}

\usepackage{url}

\usepackage{multirow}
\usepackage{arydshln} 
\usepackage{graphicx}

\aclfinalcopy % Uncomment this line for the final submission

\title{Generating Pertinent and Diversified Comments with Topic-aware Pointer-Generator Networks}

\author{
  Junheng Huang\thanks{huangjunheng@baidu.com} \\
  Baidu, Inc., China\\\And
  Lu Pan \\
  Baidu, Inc., China \\\And
  Kang Xu \\
  NJUPT, Nanjing \\\And
  Weihua Peng \\
  Baidu, Inc., China \\\And
  Fayuan Li \\
  Baidu, Inc., China \\
  }

\date{}

\begin{document}
\begin{CJK}{UTF8}{gbsn}
\maketitle
\begin{abstract}
% \footnote{email: huangjunheng@baidu.com, panlu01@baidu.com}
Comment generation, a new and challenging task in Natural Language Generation (NLG), attracts a lot of attention in recent years. However, comments generated by previous work tend to lack pertinence and diversity. In this paper, we propose a novel generation model based on Topic-aware Pointer-Generator Networks (TPGN), which can utilize the topic information hidden in the articles to guide the generation of pertinent and diversified comments. Firstly, we design a keyword-level and topic-level encoder attention mechanism to capture topic information in the articles. Next, we integrate the topic information into pointer-generator networks to guide comment generation. Experiments on a large scale of comment generation dataset show that our model produces the valuable comments and outperforms competitive baseline models significantly.

% we synthesize topic information by a LDA model and an extractive model. Next, we integrate the topic information into pointer-generator networks to guide comment generation. Experiments on a large scale of comment generation dataset show that our model produces the multiple valuable comments and outperforms competitive baseline models significantly.

% Comment generation, a new and challenging task in Natural Language Generation (NLG), attracts a lot of attention in recent years. Comments generated by previous works tend to lack informativeness and diversity. In this work, we propose a novel generation model based on Topic-aware Pointer-Generator Networks (TPGN), which focuses on constructing multiple topic-aware information in articles to guide informative and diversified comments generation. In detail, we design a method for building training dataset. Then, we propose a word-level and a topic-level encoder attention mechanisms to capture the topic-aware information in the article. Finally, we leverage the topic-aware information to guide comment generation. Experiments on a large scale comments dataset show that our model produces the multiple high-quality comments and achieves the state-of-the-art performance.
\end{abstract}

\section{Introduction}

Comments of online articles provide a form of discussion and improve user's engagement. Automatic generation of article comments has a huge value in online forums and intelligent chatbots, etc~\cite{Qin:18}. However, automatic generation of article comments is a new, challenging and not well-studied task in Natural Language Generation (NLG)~\cite{Zheng:18}, it needs to understand the meaning of articles and generate multiple valuable comments.

Machine translation models ~\cite{Xing:17} such as Sequence-to-sequence (Seq2seq) with attention ~\cite{Bahdanau:14,Cho:15} tend to generate trival samples like “\emph{I am speechless}” or “\emph{I don't know}”. In addition, for comment generation task, the models are hard to converge when we train models on articles with all comments about various topics. To generate pertinent comments, Lin et al. \shortcite{Lin:19} select the most related comment in an article, then generate one article-comment pair for training model. However, the model discards most of the article-comment pairs and are hard to generate diversified comments. 

In this paper, we focus on the generation of pertinent and diversified comments for news articles. Given a news article, we understand topics covered in the article and generate comments toward these topics. The idea is motivated by our observation on comment written by human, people often associate a news article with topics in their mind. For example, when reading a news article about “\emph{Chinese Basketball Association}”, people may think the core leader “\emph{Yao Ming}”. Based on this knowledge, they may give a pertinent comment like “\emph{we expect the institution reformation by Yao Ming}”. We consider simulating the way people write comments with topics and propose a topic-aware network to introduce the topic information as prior knowledge in comment generation. The topic words trained by LDA ~\cite{Blei:03} topic model and the keywords extracted by Textrank ~\cite{Mihalcea:04} are encoded to the topic information by our proposed encoder attention mechnism. Finally, we leverage the topic information to guide comment generation.
% Then we use the topic information to obtain the relevant context information in article by our proposed encoder attention mechanism. used as the topic information in people's mind. Finally, we leverage the topic information and the relevant context information to guide comment generation.

% Firstly, the model respectively obtains the topic words and keywords of the news article using a pre-trained LDA ~\cite{Blei:03} model and Textrank ~\cite{Mihalcea:04}. The topic words and keywords are used as the topic information in people's mind. Then we use the topic information to obtain the relevant context information in article by keyword-level and topic-level encoder attention mechanism. Finally, we integrate the topic information and the relevant context information into pointer-generator networks to guide comment generation.

\begin{figure*}[t!]
\centering
\includegraphics[width = .9\textwidth]{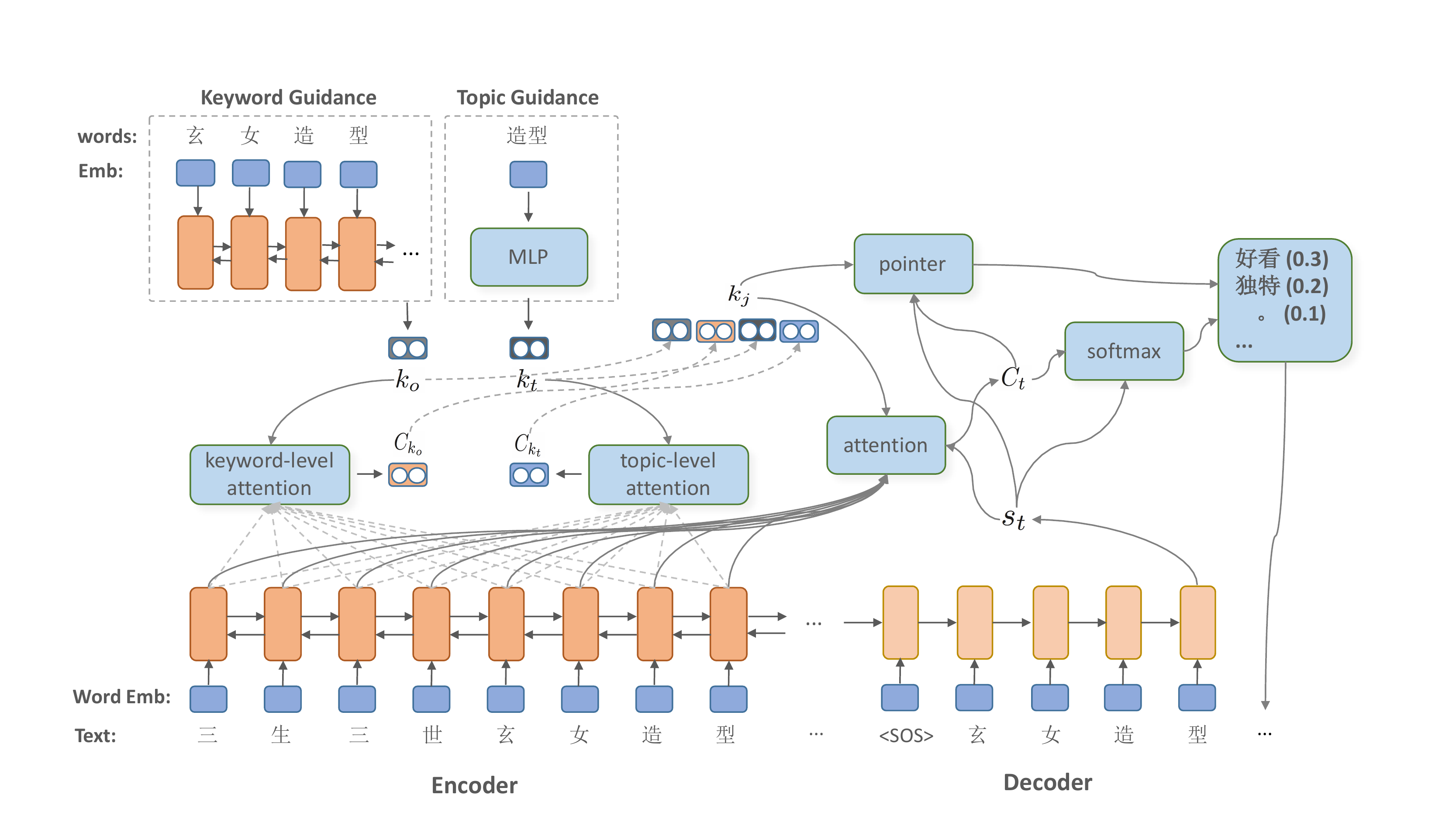}
\caption{Structure of TPGN model}
\end{figure*}

\section{Related Work}

% \subsection{Comment generation models}

{\bf Comment generation models:} the previous work ~\cite{Qin:18,Lin:19} shows that seq2seq model ~\cite{Sutskever:14}, seq2seq with attention model and pointer-generator model lead to a poor performance, when these models are trained with an article and its corresponding all comments. To improve the performance, Lin et al. \shortcite{Lin:19} propose a series of methods to select the most related comment in the article and construct one article-comment pair to train a pointer-generator model. However, the model can only generate a relevant comment for each article.

% \subsection{Prior NLG models}

{\bf Prior NLG models:} there is some work in other fields related to ours. The work ~\cite{Xing:17,Wang:18} utilize topic words embeddings learned by topic models as prior knowledge to form informative responses or summaries respectively. Li et al. \shortcite{Li:18} encodes the keywords information based on pointer-generator model to guide summarization generation. In general, our model guided by prior knowledge to produce pertinent and diversified comments.

\section{Approach}

In this section, we firstly introduce a standard pointer-generator network. Then, we summary the topic information by our keyword-level and topic-level encoder attention mechanism. Furthermore, we integrate the topic information into pointer-generator network to guide comment generation. Figure 1 gives the structure of a Topic-aware Pointer-Generator Networks (TPGN) model.

\subsection{Pointer-generator network}
Pointer-generator network is a Seq2seq-attention model with pointer network~\cite{Vin:15}. It can both copy words from input by pointer network and generate words from a fixed vocabulary. The tokens of the input article $x_i$ are fed into the encoder one-by-one, and producing a sequence of encoder hidden states $h_i$. At each decoding timestep $t$, the context vector $c_t$ is calculated by the attention mechanism ~\cite{Bahdanau:14}: 
\begin{equation}
e_{ti}=v^Ttanh(W_hh_i+W_ss_t)
\end{equation}
\begin{equation}
a_t=softmax(e_t)
\end{equation}
\begin{equation}
c_t=\sum_{n=1}^N a_{ti}h_t
\end{equation}
where $s_t$ is the decoder hidden state.
% obtained from the previous word embedding $w_{t-1}$ and context vector $c_t$. 
The generation probability $p_{gen}$ at timestep $t$ is calculated by:
\begin{equation}
p_{gen}=\sigma(W_c^Tc_t+W_s^Ts_t+W_y^Ty_{t-1}+b_{gen})
\end{equation}
where $\sigma$ is the sigmoid function. 
% Next, the vocabulary probability $P_{v}(w)$ is calculated by concatenating the context vector $c_t$ and the decoder hidden state $s_t$.
The final distribution $P(w)$ to predict the next word is calculated as following:
\begin{equation}
P(w)=p_{gen}P_{v}(w)+(1-p_{gen})\sum_{i:w_i=w}a_{ti}
\end{equation}

\subsection{Keyword-level and Topic-level encoder attention}

{\bf Keyword-level encoder attention:} we extract keywords from each article text by TextRank. The keywords for each article are fed into the BiLSTM one-by-one, then we get final hidden status $h_n$ as the keyword presentation $k_{kw}$. We use the keyword presentation $k_{kw}$ to align the article by the attention mechanism (Equation 1, 2, 3), then generate the relevant context information $C_{k_{kw}}$.

{\bf Topic-level encoder attention:} inspired by ~\cite{Xing:17}, we obtain topic embeddings by LDA topic model and use the collapsed Gibbs sampling algorithm ~\cite{heinrich:05} to estimate the parameters of the model on a dataset. After that, we select the top $n$ noun words with the highest probabilities of each topic as topic words. We calculate the embedding of each topic word by:
\begin{equation}
p(z|w)\propto\frac{C_{wz}}{\sum_{z^\prime}C_{wz^\prime}}
\end{equation}
where $C_{wz}$ is the number of times that $w$ is assigned to topic $z$. After getting the embeddings of topic words, we use a multi-layer perceptron (MLP) to obtain the core topic presentation $k_t$ for each article. From that, we convert the topic words for each article to the core topic presentation in current semantic space. Similarly, we use $k_t$ to align the article, then generate the relevant context information $C_{k_t}$. Finally, we concat $k_{kw}$, $C_{k_{kw}}$, $k_t$ and $C_{k_t}$ as the joint topic information $k_{j}$. 
\begin{equation}
k_j=[k_{kw}, C_{k_{kw}}, k_t, C_{k_t}]
\end{equation}

\subsection{Guiding comment generation}
{\bf Decoder attention:} we use the joint topic information $k_j$ as extra input to decoder attention, changing equation (1) to:
\begin{equation}
e_{ti}=v^Ttanh(W_hh_i+W_ss_t+W_kk_j)
\end{equation}
then we use the new $e_{ti}$ to obtain new attention distribution $a_{t}$ and new context vector $c_{t}$. 
% The joint topic information $k_{j}$ makes the attention mechanism pay more attention to keywords. 

{\bf Pointer mechanism:} due to the limitation of the vocabulary size, some keywords may not appear in the vocabulary, which can significantly lack the information of them in the topic information. Therefore we should encode topic information to pointer network which can copy out-of-vocabulary keywords. Similar to ~\cite{Li:18}, we use the joint topic information $k_{j}$, the context vector $c_t$ and the decoder hidden state $s_t$ as inputs to calculate $p_{gen}$, changing equation (4) to:
\begin{equation}
p_{gen}=\sigma(W_c^Tc_t+W_s^Ts_t+W_{j}^Tk_{j}+b_{gen})
\end{equation}

\begin{table*}[t!]
\centering 
\small
\label{tab:1}       % Give a unique label
\begin{tabular}{llllll}
\hline\noalign{\smallskip}
Model & TOP-N & ROUGE-L & BLEU-1 & CIDEr & METEOR \\
\noalign{\smallskip}\hline\noalign{\smallskip}
Seq2Seq-Attn & N=1 & 32.07 & 60.65 & 5.21 & 12.98 \\
pointer-generator + coverage & N=1 & 33.47 & 63.97 & 5.57 & 13.21\\
% pointer-generator + coverage + upvote & N=1 & 35.07 & 64.45 & 4.71 & 14.90\\
pointer-generator + coverage + ES & N=1 & 35.77 & 66.01 & 4.69 & 15.21\\
\cdashline{1-6}[0.8pt/2pt]
\multirow{3}{*}{KIGN} & N=1 & 40.74 & 72.21 & 11.88 & 17.98\\
& N=3 & 37.78 & 66.96 & 9.93 & 16.52\\
& N=5 & 36.67 & 64.91 & 9.18 & 15.98\\
\noalign{\smallskip}\hline\noalign{\smallskip}
\multirow{3}{*}{Keyword-level TPGN} & N=1 & 41.45 & 73.25 & 13.17 & 18.56\\
& N=3 & 37.97 & 66.63 & 10.30 & 16.82\\
& N=5 & 36.43 & 63.65 & 9.18 & 16.09\\
\cdashline{1-6}[0.8pt/2pt]
\multirow{3}{*}{Topic-level TPGN} & N=1 & 42.01 & 75.10 & 13.39 & 18.61\\
& N=3 & 38.08 & 67.41 & 10.13 & 16.56\\
& N=5 & 36.01 & 63.48 & 8.61 & 15.52\\
% \noalign{\smallskip}\hline\noalign{\smallskip} 
\cdashline{1-6}[0.8pt/2pt]
\multirow{3}{*}{TPGN} & N=1 & \textbf{45.16} & \textbf{81.47} & \textbf{16.64} & \textbf{20.73}\\
& N=3 & 40.88 & 74.01 & 12.98 & 18.39\\
& N=5 & 38.34 & 69.48 & 10.87 & 17.05\\
\noalign{\smallskip}\hline
\end{tabular}
\caption{Results on generation models. Higher scores are better.}
\end{table*}

\section{Evaluation}

\subsection{Dataset and evaluation metrics}
We obtain a large-scale Chinese article commenting dataset ~\cite{Qin:18} which has 174886 articles with four million users comments, and each article has a title, text body and metadata. The dataset is split into training/validation/test sets which contains 169023/1400/4463 samples respectively. 
For the evalution of models, we use the same metrics ~\cite{Lin:19}, and get the script from Coco Caption ~\cite{Chen:15}.
% \footnote{https://github.com/tylin/coco-caption/}

\subsection{Evaluation models}
We consider four different baselines and three variations of our propose approach. \emph{Seq2Seq-Attn:} the Seq2Seq model with attention, we train the model with each article and its all comments; \emph{pointer-generator + coverage:} the pointer-generator network ~\cite{See:17}, we construct training pairs like Seq2Seq-Attn; \emph{pointer-generator + coverage + ES:} in this model ~\cite{Lin:19}, each article only corresponds to a comment selected by the ensemble score; \emph{KIGN:} a guide network ~\cite{Li:18} is the simplified version of our approach; \emph{Keyword-level TPGN:} our approach which employs only the keyword-level encoder attention; \emph{Topic-level TPGN:} our approach which employs only the topic-level encoder attention; \emph{TPGN:} our approach which employs both the keyword-level and the topic-level encoder attentions.

% {\bf pointer-generator + coverage + upvote:} the model ~\cite{Lin:19} is the same as ~\cite{See:17}. The difference is that each article only corresponding to a article-comment pair, the comment is selected from the highest upvote comment.

\begin{table*}[]
\centering
% \small
\footnotesize
\begin{tabular}{|l|l|l|l|}
\hline
\begin{tabular}[c]{@{}l@{}}Title: 刘亦菲杨洋《三生三世》\\ 玄女造型“辣眼睛”\\ (The costume of Xuan Nv \\"stings eyes" in \textless{}\textless{}Three \\ Lifetimes\textgreater{}\textgreater where Liu Yifei \\ and Yang Yang played the roles)\end{tabular} & Model & \begin{tabular}[c]{@{}l@{}}Keywords\\ and Topic words\end{tabular} & Generated Comment \\ \hline
\multirow{6}{*}{\begin{tabular}[c]{@{}l@{}}Body: 电影版《三生三世》上\\映，票房一路高歌猛进，虽然\\不敌 《战狼2》，但票房也不\\错。而剧中最“辣眼睛”的造型\\......罗晋饰演折颜， \\他的服装和妆容也被指辣\\眼睛，完全搭配得像“凤凰”。\\ (The movie version of \textless{}\textless{}Three \\ Lifetimes\textgreater{}\textgreater comes out. His box \\ office is rapidly increasing, but \\ less than \textless{}\textless{}Wolf Warriors 2\textgreater{}\textgreater\\  ...... Luo Jin plays the role called\\ Zhe Yan, whose costume is \\ pointed to "sting eyes", just \\ like the "phoenix")\end{tabular}} & \begin{tabular}[c]{@{}l@{}}pointer-generator \\ + coverage\end{tabular} & - & \begin{tabular}[c]{@{}l@{}}我也是醉了。。。。。\\ (I am speechless . . . . .)\end{tabular} \\ \cline{2-4} 
 & \begin{tabular}[c]{@{}l@{}}pointer-generator \\ + coverage + ES\end{tabular} & - & \begin{tabular}[c]{@{}l@{}}刘亦菲最帅。。。。。\\ (Liu Yifei is the most handsome.....)\end{tabular} \\ \cline{2-4} 
 & \multirow{2}{*}{KIGN} & \begin{tabular}[c]{@{}l@{}}杨洋，刘亦菲\\ (Yang Yang, \\ Liu Yifei )\end{tabular} & \begin{tabular}[c]{@{}l@{}}我觉得刘亦菲演的很好看。\\ (I think Liu Yifei is good at acting.)\end{tabular} \\ \cline{3-4} 
 &  & \begin{tabular}[c]{@{}l@{}}罗晋，服装\\ (Luo Jin, Costume)\end{tabular} & \begin{tabular}[c]{@{}l@{}}罗晋最帅！！！！！！\\ (Luo Jin is the most handsome......)\end{tabular} \\ \cline{2-4} 
 & \multirow{2}{*}{TPGN} & \begin{tabular}[c]{@{}l@{}}杨洋，\textbf{刘亦菲}\\ (Yang Yang, \\ \textbf{Liu Yifei} )\end{tabular} & \begin{tabular}[c]{@{}l@{}}刘亦菲杨洋版本本来就不错。\\ (The movie version of Liu Yifei \\ and Yang Yang are supposed\\ to be really good.)\end{tabular} \\ \cline{3-4} 
 &  & \begin{tabular}[c]{@{}l@{}}罗晋，\textbf{服装}\\ (Luo Jin, \textbf{Costume})\end{tabular} & \begin{tabular}[c]{@{}l@{}}玄女好看，我喜欢罗晋的服装搭配。\\ (Xuan Nv is beautiful, I like \\the costume of Luo Jin. )\end{tabular} \\ \hline
\end{tabular}
\caption{The generated comments from different generation models (\textbf{bold} denotes Topic words).}
\end{table*}

\subsection{Experimental setting}
For all experiments, we use two 256-dimensional LSTMs for the encoder and one 256-dimensional LSTM for the decoder. We use a word embedding with a size of 128 with a vocabulary size of 9k. We train models by using Adagrad ~\cite{Duchi:11} with learning rate 0.1 and an initial accumulator value of 0.1. 
% We truncate the article to 400 characters and the comment to 50 characters. 
For KIGN model and our approach, we extract keywords from each article by TextRank. Comments in the article can be matched by the combinations of different keywords. Then, we obtain a series of triple (article, keywords, matched comment) for each article. If can't find a matched comment for the article, we randomly choose a comment under this article. We use a LDA model to achieve the embeddings of topic words, and the number of topics $T$ is set to 100. 
For each topic, we select the top 500 words as topic words. If can't find corresponding topic words for the article, we obtain the uniform distribution on the dimension of the number of topics.

\subsection{Results and analysis}

For each article, our models can produce multiple comments according to different keywords\footnote{In predict stage, we select keywords by Textrank from every sentence in the article}, so we also calculate top N highest scores. Table 1 shows that KIGN model achieves the best scores in baselines, the models of our approach all outscore the best baseline model while N in (1, 3, 5). In average, KIGN, Keyword-level TPGN, Topic-level TPGN and TPGN model respectively generate 2.9, 5.1, 4.8 and 6.2 different comments for each article after removing duplicates. It means that almost comments generated by our models are used to calculate evaluation scores. In addition, TPGN model achieves the best performance while N equals 1, and exceeds KIGN model by (+4.4 ROUGE-L, +9.2 BLEU-1, +4.8 CIDEr, +2.8 METEOR). Table 2 compares our model with three baselines using an example. We find that pointer-generator + coverage model generates a trival comment such as “I am speechless”. Pointer-generator + coverage + ES model captures the key information of the article, but produces a repetitive and uninteresting comment. KIGN model generates two useful comments according to the different combinations of keywords. Moreover, our TPGN model, which associates with the topic information, generates more pertinent and diversified comments.

\section{Conclusion}

% In this work, we propose a guiding generation model for comment generation. Firstly, we design a word-level and topic-level encoder attention mechanism to capture topic information in the articles. Next, we integrate the topic information into Pointer-Generator Networks to guide pertinent and diversified comment generation. Experiments on a large scale of comments generation dataset show that our model produces the multiple high-quality comments and achieves the state-of-the-art performance. For future work, we plan to construct a hierarchical structure to learn article presentation and combine it with keyword information to generate a more related guiding information. 

In this work, we propose a topic-aware generation model for comment generation. Firstly, we design a encoder attention mechanism to capture the topic information in the articles. Then, we leverage the topic information to guide comment generation. Experiments show that our model produces the pertinent and diversified comments and achieves the state-of-the-art performance. 
% Min: no longer used as of ACL 2018, following ACL exec's decision to
% remove this extra workflow that was not executed much.
% BEGIN: remove
%% \section{XML conversion and supported \LaTeX\ packages}

%% Following ACL 2014 we will also we will attempt to automatically convert 
%% your \LaTeX\ source files to publish papers in machine-readable 
%% XML with semantic markup in the ACL Anthology, in addition to the 
%% traditional PDF format.  This will allow us to create, over the next 
%% few years, a growing corpus of scientific text for our own future research, 
%% and picks up on recent initiatives on converting ACL papers from earlier 
%% years to XML. 

%% We encourage you to submit a ZIP file of your \LaTeX\ sources along
%% with the camera-ready version of your paper. We will then convert them
%% to XML automatically, using the LaTeXML tool
%% (\url{http://dlmf.nist.gov/LaTeXML}). LaTeXML has \emph{bindings} for
%% a number of \LaTeX\ packages, including the ACL 2018 stylefile. These
%% bindings allow LaTeXML to render the commands from these packages
%% correctly in XML. For best results, we encourage you to use the
%% packages that are officially supported by LaTeXML, listed at
%% \url{http://dlmf.nist.gov/LaTeXML/manual/included.bindings}
% END: remove

\bibliography{emnlp-ijcnlp-2019}
\bibliographystyle{acl_natbib}
\end{CJK}
\end{document}